\documentclass[11pt]{article}
\usepackage[utf8]{inputenc}
 
\usepackage[final]{acl}
 
\usepackage{times}
\usepackage{latexsym}
 
\usepackage[T1]{fontenc}

\usepackage{microtype}
 
\usepackage{inconsolata}
 
\usepackage{graphicx}

\usepackage{times}
\usepackage{latexsym}
\usepackage{amsmath}
\usepackage{booktabs}
\usepackage{pifont}
\usepackage{graphicx}
\usepackage{listings}
\usepackage{amssymb}
\usepackage{multirow}
\usepackage[T1]{fontenc}
\usepackage{microtype}
\usepackage{inconsolata}
\usepackage{xcolor}
\usepackage{upquote} 

\usepackage[most]{tcolorbox}
\tcbuselibrary{listings, breakable}

\newtcblisting{promptbox}[1][]{
  breakable, enhanced,
  colback=gray!5, colframe=gray!45, boxrule=0.4pt, arc=2pt,
  left=4pt, right=4pt, top=4pt, bottom=4pt,
  fonttitle=\small\bfseries, coltitle=black, colbacktitle=gray!15,
  listing only,
  listing options={basicstyle=\footnotesize\ttfamily, breaklines=true,
                   columns=flexible, keepspaces=true},
  #1
}

\title{RelayS2S: Dual-Path Speculative Generation for Real-Time Dialogue}
 
\author{
  Long Mai \\
  Trinity College Dublin, Ireland \\
  \texttt{mailong225@gmail.com}
  \And
  Junli Liang \\
  University College Dublin, Ireland \\
  \texttt{junli.liang@ucd.ie}
}
 
\begin{document}
\maketitle
 
\begin{abstract}
Real-time spoken dialogue systems face a fundamental tension between latency and response quality. End-to-end speech-to-speech (S2S) models respond immediately and naturally handle turn-taking, backchanneling, and interruption, but produce semantically weaker outputs. Cascaded pipelines (ASR $\rightarrow$ LLM) deliver stronger responses at the cost of latency that grows with model size. We present RelayS2S, a hybrid architecture that runs two paths in parallel upon turn detection. The fast path --- a duplex S2S model --- speculatively drafts a short response prefix that is streamed immediately to TTS for low-latency response onset, while continuing to monitor live audio events. The slow path --- a cascaded ASR $\rightarrow$ LLM pipeline --- generates a higher-quality continuation conditioned on the committed prefix, producing an uninterrupted utterance. A lightweight learned verifier gates the handoff, committing the prefix when appropriate or falling back gracefully to the cascaded pipeline. With GPT-4.1 as the back-end, RelayS2S substantially reduces response latency while preserving nearly all of the cascaded pipeline's textual quality. On synthetic voice dialogues, it achieves a P90 first-chunk latency of 81\,ms, excluding TTS and network latency, compared with 1{,}006\,ms for the cascaded baseline. On real voice dialogues, RelayS2S reduces average first-chunk latency by 479\,ms while retaining 99\% of the cascaded pipeline's textual quality. These benefits become larger as the slow-path model scales. Because the prefix handoff requires no architectural modification to either component, RelayS2S serves as a lightweight, drop-in addition to existing cascaded pipelines. Our code is publicly available at \url{https://github.com/mailong25/relays2s}.
\end{abstract}

\section{Introduction}
 
Real-time spoken dialogue is among the most demanding settings for conversational AI: a system must listen, reason, and respond with the speed and fluidity of human conversation. Existing approaches face a central trade-off. End-to-end full-duplex speech-to-speech models~\cite{zhang2023speechgpt, rubenstein2023audiopalm, defossez2024moshi, nguyen2025salm, li2025personaplex} can begin speaking almost immediately and natively support other turn-taking behaviors, but produce semantically weaker outputs. Cascaded pipelines (ASR $\rightarrow$ LLM)~\cite{chen2025fireredchat, liu2025xtalk, faster_whisper, likhomanenko2025chipchat} deliver stronger responses, but their onset latency often exceeds the 200\,ms response gap typical of human conversations \cite{levinson2015timing}.

We build on a simple empirical observation: although S2S responses are weaker overall, their first few words are frequently usable. In our analysis, 82.5\%--95\% of five-word S2S prefixes are judged contextually appropriate---high enough to make speculative prefixing viable when paired with a learned verifier. Motivated by this asymmetry, we propose \textbf{RelayS2S}, a speculative hybrid architecture that runs two paths in parallel upon turn detection. The \textit{fast path}---a duplex S2S model---drafts a short response prefix streamed immediately to TTS. The \textit{slow path}---a cascaded ASR--LLM pipeline---generates a higher-quality continuation conditioned on the committed prefix. A lightweight learned verifier gates the handoff, committing the prefix when appropriate or falling back to the slow path alone.
 
A key design challenge is that the fast-path model must generate a prefix and continue monitoring live audio for interruptions. We resolve this with \textit{forked speculative generation}: upon deciding to speak, the model forks into a main stream that tracks live audio for barge-in detection and a speculative stream that drafts the prefix at maximum speed. Unlike concurrent approaches that emit only short discourse connectives \cite{liu2026ddtsr}, rely on text-only pipelines that delay audible response despite early reasoning \cite{zou2026lts}, or inject oracle LLM tokens into S2S transformers via joint training \cite{kuroki2025kame}, RelayS2S generates substantive spoken prefix content while natively supporting full-duplex conversational behaviors, and the prefix handoff requires no architectural modification to either component.

We train the fast-path on 194K synthetic and real voice conversations (2{,}437 hours) and evaluate across four back-ends. With GPT-4.1 \cite{gpt41} as the back-end, RelayS2S achieves a P90 first-chunk latency of 81\,ms on synthetic voice dialogues---down from 1{,}006\,ms for the cascaded baseline---while preserving 99\% of the cascaded system's average quality score. On the real voice dialogues, RelayS2S cuts average first-chunk latency by 479\,ms (269\,ms vs.\ 748\,ms) at comparable quality retention. Our main contributions are:
\begin{itemize}
    \item \textbf{Forked speculative generation}, which decouples fast response drafting from real-time audio monitoring within a single duplex S2S model, enabling immediate prefix generation. The resulting buffered multi-token prefix  serves two downstream roles in the architecture: it provides the prefix verifier with a meaningful sequence of hidden states and calibration signals over which to score the draft, and it lets streaming TTS render from longer text segments, which generally improves synthesis quality \cite{speakstream2025}.
    
    \item \textbf{Selective prefix handoff}, in which a lightweight learned verifier gates whether a speculative prefix is safe to commit. The verifier reuses already-computed decoder hidden states and calibration signals, adding negligible overhead while achieving 95\% good-prefix throughput at an 8\% fallback rate on synthetic dialogue.
    
    \item \textbf{RelayS2S}, a hybrid pipeline that combines a duplex S2S model with a cascaded back-end and scales monotonically with progress in either subsystem, cutting response latency substantially while preserving nearly all of the slow path's textual quality across four LLMs.
\end{itemize}
 
\section{RelayS2S}
\label{sec:system}
 
\begin{figure*}[t]
    \centering
    \includegraphics[width=\textwidth]{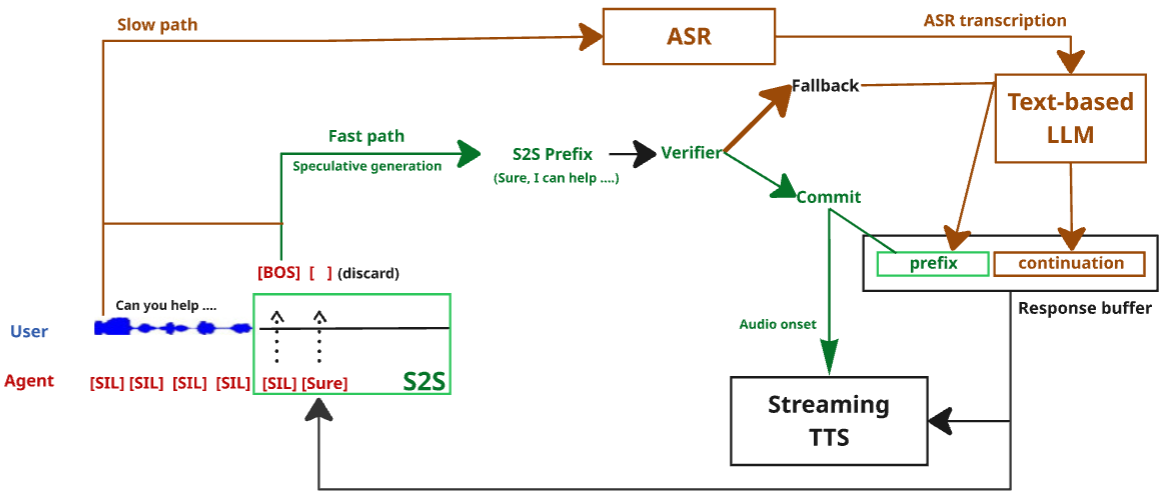}
    \caption{Inference-time architecture of RelayS2S, including the fast path (green) and the slow path (brown).}
    \label{fig:relays2s}
\end{figure*}

\subsection{Overview}
\label{sec:overview}
 
Figure~\ref{fig:relays2s} illustrates the inference-time design. Upon
detecting the end of the user's turn, RelayS2S launches two parallel paths.
The fast path (\S\ref{sec:fast_path}) uses an S2S model to draft the opening
few words of the response. A verifier (\S\ref{sec:gating}) decides whether to
commit the draft to streaming TTS. If committed, the slow path
(\S\ref{sec:slow_path})---a cascaded ASR--LLM pipeline---generates the
remainder conditioned on the committed prefix; otherwise it generates the
full response and the system behaves like a conventional cascade. The
handoff between the two paths occurs entirely in the text domain: both write
to a shared response buffer consumed by a single streaming TTS session, so
the prefix and its continuation are rendered by the same voice and session
rather than spliced from two independently synthesized audio segments.
Because a 5-word prefix yields roughly 2 seconds of speech, it creates a
temporal buffer (the \emph{relay margin}; \S\ref{sec:tts}) that lets the slow
path produce its first continuation chunk before the prefix audio is
exhausted, yielding a single uninterrupted utterance. While the prefix is
playing, the fast-path continues to monitor user speech so that any
interruption stops playback immediately.

\subsection{Fast Path: Duplex S2S Model}
\label{sec:fast_path}
\begin{figure}[t]
    \centering
    \includegraphics[width=0.47\textwidth]{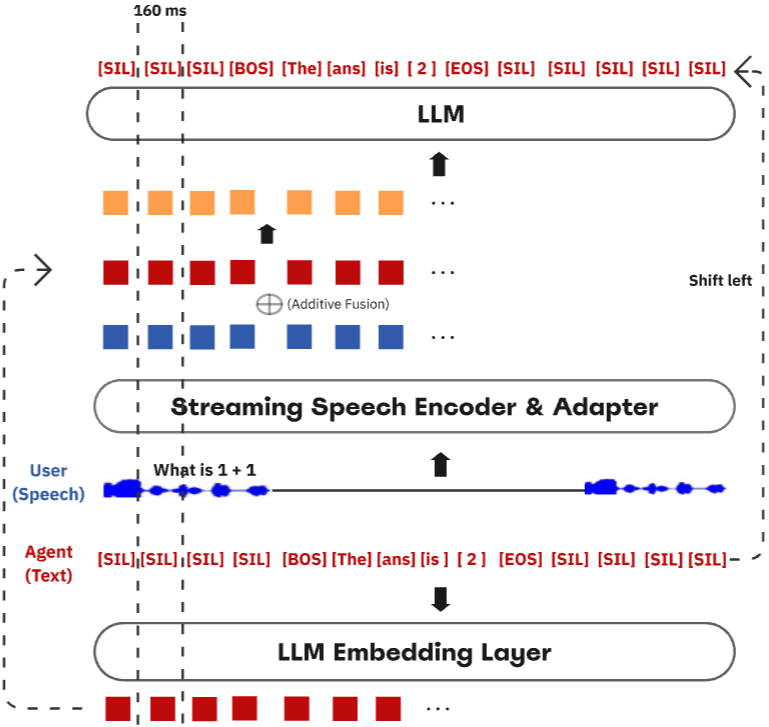}
    \caption{Fast-path duplex S2S model}
    \label{fig:s2s_arch}
\end{figure}

The fast path is a duplex speech-conditioned response model that maps streaming user speech directly to agent-side control symbols and text response tokens, which are then rendered by streaming TTS. It serves two roles in RelayS2S: enabling low-latency conversational behavior (early response initiation, backchanneling, interruption handling) and drafting an initial response prefix for handoff to the slow path. It comprises a streaming speech encoder, an LLM adapter, and an autoregressive LLM backbone (Figure~\ref{fig:s2s_arch}).

\paragraph{Streaming speech encoder.}
We adopt the encoder of \citet{forzen_llm}: a conformer with chunked
attention and limited left context that consumes audio causally. Log-Mel
filterbank features are processed into higher-level acoustic
representations, with internal subsampling producing a compact frame
rate. Full hyperparameters are given in Appendix~\ref{app:implementation}.

\paragraph{LLM adapter.}
A two-stage causal 1D convolutional stack (convolution, normalization,
GELU per stage) further compresses the temporal resolution and projects
encoder states into the LLM embedding space. The combined encoder--adapter
emits one representation every 160\,ms, defining the online inference
update interval.

\paragraph{LLM backbone.}
An autoregressive LLM consumes the adapted representations. Its
vocabulary is extended with duplex control symbols:
\texttt{[SIL]} (continue listening),
\texttt{[BOC]} (begin backchannel),
\texttt{[BOS]} (begin substantive response),
\texttt{[STP]} (terminate ongoing response, e.g., on interruption), and
\texttt{[EOS]} (complete response).
Ordinary text tokens carry response content, letting a single decoder
unify dialogue control and lexical generation. At each step, the user
speech representation $\mathbf{e}^{\text{speech}}_t$ is fused with the
aligned agent token embedding $\mathbf{e}^{\text{text}}_t$ by element-wise
addition and passed through the LLM to predict the next agent token.
 
\subsection{Forked Speculative Generation}
\label{sec:forked}
 
When the fast-path model decides to initiate a response by emitting
\texttt{[BOS]} or \texttt{[BOC]}, it must generate a text prefix quickly
enough for streaming TTS to begin synthesis with minimal delay. A naive
approach---continuing to emit one token per 160\,ms update tick and forwarding
tokens incrementally to TTS---is too slow for natural interaction.
High-quality streaming TTS typically requires a short buffered chunk of text
before it can begin rendering~\cite{speakstream2025}. If the minimum chunk
size is $C$ words, the earliest audio onset under tick-synchronous generation
would be delayed by at least $C \times 160$\,ms. For realistic chunk sizes of
5--8 words, this corresponds to roughly 0.8--1.3\,s, which exceeds the
threshold for natural conversational flow. To avoid this bottleneck, we \textit{fork} the fast path after a response
initiation event into two branches: a \textbf{main online stream} and a
\textbf{speculative generation stream}.
 
\paragraph{Main online stream.}
This branch remains the authoritative execution path. It continues processing
live user speech every 160\,ms and maintains the primary KV-cache, preserving
the model's ability to detect barge-in and other real-time interaction events.
The main stream does not generate the response tokens itself but consumes the tokens produced by the speculative stream as
agent-side history, keeping its autoregressive state aligned with the response
that the listener is already hearing.
 
\paragraph{Speculative stream.}
This branch is initialized from the same decoder state at timestep
$t_{\text{init}}$ but stops observing future user speech inputs, which in our
implementation is realized by setting
\begin{equation}
    \mathbf{e}_t^{\text{speech}} = \mathbf{0}, \qquad \forall\, t \ge t_{\text{init}}
\end{equation}
Freed from the 160\,ms speech-tick constraint, the speculative branch switches
to free-running autoregressive decoding and generates a short text prefix at
the model's maximum decoding speed. If approved by the prefix verifier
(\S\ref{sec:gating}), these drafted tokens are buffered and forwarded to the
streaming TTS system as soon as the minimum synthesis chunk is reached,
substantially reducing the time to first audio.
 
\paragraph{Interruption handling.}
Because the speculative stream no longer observes live user speech, it cannot detect
barge-in or interruptions from user. The main online stream retains this responsibility. If the main
stream predicts \texttt{[STP]}, playback is halted immediately and the
remaining buffered draft is discarded. This design decouples fast response generation from real-time interaction monitoring, achieving low response-onset latency without sacrificing interruption handling.
 
\subsection{Selective Prefix Handoff}
\label{sec:gating}
Once the speculative stream drafts a prefix, a verifier decides whether to commit it for immediate streaming to TTS or suppress it and defer entirely to the cascaded pipeline. When committed, the slow path continues generation conditioned on the prefix. Because the verifier reuses already-computed decoder states, it adds negligible overhead.

\paragraph{Verifier input representation.}
At each drafted token position $t \in \{1, \dots, n\}$, we extract two
signals: (i)~the final-layer decoder hidden state
$\mathbf{h}_t \in \mathbb{R}^{d}$, capturing contextual semantics, and
(ii)~a scalar feature vector $\mathbf{s}_t \in \mathbb{R}^{3}$ comprising the
output distribution entropy $H(P_t)$, the selected-token log-probability
$\log P_t(y_t)$, and the top-two logit margin
$z_t^{(1)} - z_t^{(2)}$, where $z_t^{(1)}$ and $z_t^{(2)}$ are the
largest and second-largest pre-softmax logits at step $t$. Under greedy
decoding this margin coincides with the log-probability gap between the
top two candidates. Together, these signals provide both semantic
and calibration information per token.
 
\paragraph{Verifier architecture.}
The verifier projects each hidden state to a lower-dimensional space,
modulates it with the scalar features via a multiplicative gate plus an
additive bypass, adds learned positional embeddings, and applies a
token-level feedforward block with a residual connection. The resulting
per-position representations are aggregated by attention pooling with a
single learned query and passed through a sigmoid head to produce a
confidence score $c \in [0, 1]$. Full architectural equations and
dimensions are provided in Appendix~\ref{app:verifier_arch}.
 
\paragraph{Training.}
The verifier is trained as a binary classifier over drafted prefixes labeled as \texttt{good} or \texttt{bad}. We generate speculative prefixes by running the S2S model on its own training split, then produce a prefix-conditioned continuation with GPT-5.4 \cite{gpt5} and submit the (context, prefix, continuation) triple to an ensemble of three LLMs whose majority vote sets the final label (see Appendix~\ref{app:prefix_check} for prompts and LLM
details). A prefix is labeled \texttt{good} if it is judged contextually sensible given the dialogue context, and \texttt{bad} otherwise. The validation and test sets for the prefix verifier are obtained from the corresponding validation and test splits of the S2S model. To ensure high-fidelity evaluation signals, the ground-truth labels for both the validation and test sets are annotated by human annotators.
 
\subsection{Slow Path: Cascaded Pipeline}
\label{sec:slow_path}
 
The slow path is a conventional ASR--LLM pipeline whose role is to produce a higher-quality response in text space once the fast path has secured low-latency response onset.
 
\paragraph{Triggering and ASR.}
Rather than relying on a separate end-of-utterance detector, the slow path
reuses the fast path's \texttt{[BOS]} decision as its trigger. Because this
decision already implies that the user has yielded the floor, the full
utterance is typically available at trigger time. The buffered speech segment is
passed to an ASR model, and the resulting transcript is combined with
the dialogue history to form the LLM input context.
 
\paragraph{Prefix-conditioned generation and self-correction.}
If the verifier commits the fast-path prefix, the committed tokens are
supplied to the slow-path LLM as a hard constraint---so that the model's output is conditioned on the exact token sequence the listener is already hearing. For local LLMs, we prefill the assistant message with the committed prefix and continue decoding from that state. For API LLMs, the prefix is passed as already-spoken assistant context with instructions to emit only the continuation.

The slow-path LLM is also prompted to handle prefix failures
dynamically. When the prefix is contextually sound, the model
continues it smoothly. When the prefix contains an early factual
error, hallucination, or contextual inconsistency, the slow path
executes a conversational repair within the continuation stream
(\textit{e.g.}, \textit{``...wait, I mean...''} or \textit{``...actually,
let me correct that...''}) rather than propagating the flaw. The full
prompt template appears in Appendix~\ref{app:response_continuation}.
 
\subsection{Streaming TTS}
\label{sec:tts}
The streaming TTS module ~\cite{elevenlabs_flash_v25} synthesizes
chunk-incrementally, beginning as soon as a minimum text chunk is
available. For uninterrupted audio, the slow-path continuation must arrive
before the TTS exhausts the prefix; we call this gap the
\textit{relay margin}, with positive values yielding an uninterrupted
utterance and negative ones producing a perceptible pause. In
practice, a 5-word prefix yields roughly 2 seconds of speech,
comfortably exceeding the slow-path generation time for most LLM back-ends.

\section{Training Data}
\label{sec:data}
 
We train on a 2{,}437-hour, 194K-conversation corpus combining
(a)~104K fully synthetic dialogues (2{,}133\,h) synthesized with
CosyVoice2~\cite{cosyvoice} and noise-augmented from
TAU~\cite{tau_noise}, and (b)~90K real-voice two-turn conversations
(304\,h) built by pairing filtered human audio from
GigaSpeech~\cite{gigaspeech}, Common Voice~\cite{common}, and
Switchboard~\cite{switchboard} with GPT-5.4-generated
assistant responses. We additionally inject backchannels, interruptions,
and mid-utterance pauses to train the duplex control tokens described in
\S\ref{sec:fast_path}. Full data construction details are
provided in Appendix~\ref{app:training_data}.
 
\section{Experiments}
\label{sec:experiments}
 
\subsection{Implementation Details}
\label{sec:implementation}
 
The fast-path S2S model uses a streaming conformer encoder, a two-stage causal CNN adapter producing one speech representation every 160\,ms, and a Qwen2.5-0.5B~\cite{qwen2.5} LLM backbone. Training proceeds in three stages: (1)~encoder--adapter pretraining on speech transcription; (2)~full-duplex S2S training end-to-end on the data described in \S\ref{sec:data}; and (3)~prefix verifier training as described in \S\ref{sec:gating}. More training details are provided in Appendix~\ref{app:implementation}.
 
\subsection{Baselines and Concurrent Works}
\label{sec:baselines}
 
We compare RelayS2S against two primary baseline families:
 
\paragraph{S2S only.}
This baseline uses the same fast-path model with forked free-running generation. It is functionally similar to native end-to-end full-duplex speech architectures such as \mbox{SALM-Duplex} \cite{nguyen2025salm}, but without a slow-path LLM or prefix verifier. This setup isolates the performance of the fast path without any cascaded guidance.

\paragraph{Cascaded ASR--LLM.}
We pair an offline ASR module with text-based LLMs of varying capacity.
For ASR, we use Whisper~\cite{faster_whisper} (\texttt{medium.en})
via the CTranslate2 implementation\footnote{https://github.com/SYSTRAN/faster-whisper}. The model transcribes a user
turn in approximately 200\,ms on average using NVIDIA L40S, comparable to the end-of-turn finalization latency of production streaming ASR systems. For the text-based LLMs, we evaluate four
configurations spanning both local (Qwen2.5-0.5B, Qwen2.5-7B) and
API-based inference (Grok-4.1-fast~\cite{grok41}) and
GPT-4.1. We choose this sequential pipeline to prioritize response quality. Although incremental systems \cite{pred_gen} can reduce latency by generating from partial ASR hypotheses, later revisions or late-arriving user constraints may require rollback and degrade response quality. Nevertheless, incremental inference is complementary to our approach and could be incorporated into the slow path to further reduce fallback latency.

We also compare against two recent concurrent works, DDTSR \citet{liu2026ddtsr} and KAME \citet{kuroki2025kame}. For
each, we isolate the architectural axis the system explores by evaluating
the corresponding RelayS2S configuration.

\textbf{DDTSR} proposes a dual-track architecture in
which the fast path emits only short discourse connectives (1--3 tokens)
while the back-end LLM reasons in parallel. Public code, models, and
training data remain unavailable at the time of writing, so we approximate
this regime with RelayS2S at \texttt{prefix\_length=3}, isolating the
effect of a short speculative prefix in a conventional streaming pipeline
(Tables~\ref{tab:relay_margin} and~\ref{tab:ablation}).

\textbf{KAME} uses a tandem architecture in which an
S2S front-end speaks first and is progressively corrected by an
asynchronous LLM ``oracle'' stream, with no gating between drafted and
committed tokens. We characterise this design along two complementary
axes. First, to isolate un-gated commitment within a fixed architecture, we
disable the prefix verifier in RelayS2S so that every drafted prefix is
committed unconditionally (Table~\ref{tab:ablation}). Second, we evaluate
the publicly released KAME checkpoint directly\footnote{\url{https://github.com/SakanaAI/kame}}. Because that checkpoint is trained on a different data
distribution and targets reasoning benchmarks rather than open-ended
conversation, we report it in Appendix~\ref{app:kame} as a
released-checkpoint evaluation rather than as a controlled architectural
comparison.

\subsection{Evaluation Metrics}
\label{sec:metrics}
 
\paragraph{Textual response quality.}
We prompt Gemini-3-Flash \cite{gemini3flash} to rate each generated response on a 1--5 scale given the
dialogue context. Because absolute scores can be subjective and exhibit
non-trivial variance across runs, we report both the average score and the
\textit{low-quality rate}: the fraction of responses scored $\leq 3$, where a
score of 3 or below indicates noticeable issues under our rubric
(Appendix~\ref{app:textual_quality}). We validate the Gemini-3-Flash judge
against a 3-annotator human panel (Cohen's $\kappa = 0.70$ vs.\ majority vote)
and verify run-to-run stability ($\sigma = 0.47$\,pp on low-quality rate,
Fleiss' $\kappa = 0.81$); full details in Appendix~\ref{app:judge_validation}.
We adopt the low-quality rate as our primary headline metric because it offers
superior interpretability (the percentage of noticeably problematic responses)
and is robust to minor calibration shifts in the 4--5 range that do not affect
the practical acceptability boundary. We report 95\% bootstrap percentile confidence intervals ($B = 5{,}000$ resamples, seed $= 42$) for all quality metrics.
 
\paragraph{First-chunk latency.}
To isolate the core algorithmic efficiency of our architecture, we report first-chunk latency: the interval from turn detection (the moment the agent decides to speak) to the availability of the agent's first $N$ response words for the streaming TTS module.
We adopt $N = 5$ as our default operating point, following \citet{speakstream2025}, who identify a text window length of 5 as optimal for high-quality streaming speech synthesis. Also, a 5-word chunk yields roughly 2 seconds of speech, providing a sufficient temporal buffer for the slow path to deliver its first continuation chunk. We explicitly exclude infrastructure-dependent variables—specifically raw turn-detection lag and TTS audio rendering time—as these components are shared identically across all baselines and do not affect relative evaluation. Under this definition, first-chunk latency is computed as follows:

\noindent \textbf{S2S only:} time from turn detection to the S2S decoding of
the $N$-th response word: $L_{\text{S2S}}$. \\
\textbf{Cascaded only:} ASR transcription time plus LLM generation of
the first $N$ response words:
$L_{\text{cascaded}} = T_{\text{ASR}} + T_{\text{LLM}}$. \\
\textbf{RelayS2S (\textsc{commit}):} S2S decoding time plus the
verifier decision:
$L_{\text{commit}} = L_{\text{S2S}} + T_{\text{verifier}}$. \\
\textbf{RelayS2S (\textsc{fallback}):} falls back to the cascaded
path: $L_{\text{fallback}} = L_{\text{cascaded}}$.

\noindent Latency is measured on a single NVIDIA L40S GPU for local
models, and client-side (dispatch to first chunk) for API back-ends.

\subsection{Results}
\label{sec:results}
 
We evaluate our system on both synthetic and real dialogue test sets. The 
synthetic subset comprises 1,000 held-out conversations (2,175 
context--response pairs), while the real subset contains 2,000 held-out 
conversations (2,000 context--response pairs). For each 
configuration, we report the average and P90 first-chunk latency, alongside the textual response quality. Table~\ref{tab:results} 
summarizes the performance across these subsets. Note that the fast-path 
component remains identical across all evaluated RelayS2S configurations, 
utilizing the same 0.5B S2S model and prefix verifier; performance variations 
are driven solely by the choice of the slow-path LLMs.

\begin{table*}[t]
\centering
\small
\renewcommand{\arraystretch}{1.15}
\setlength{\tabcolsep}{5pt}
\begin{tabular}{llcccc}
\toprule
\textbf{System} & \textbf{Slow-path LLM} & \textbf{Avg Lat.\ (ms)} & \textbf{P90 Lat.\ (ms)} & \textbf{Avg Quality} & \textbf{Low-quality (\%)} \\
\midrule
\multicolumn{6}{l}{\textbf{Synthetic voice dialogues}} \\
S2S only         & ---            &  59 &   72 & $3.29_{\pm 0.06}$ & $52.7_{\pm 2.1}$ \\
\midrule
Cascaded         & Qwen2.5-0.5B   & 304 &  418 & $3.45_{\pm 0.06}$ & $48.7_{\pm 2.1}$ \\
RelayS2S         & Qwen2.5-0.5B   &  77 &   81 & $3.49_{\pm 0.06}$ & $47.6_{\pm 2.1}$ \\
\midrule
Cascaded         & Qwen2.5-7B     & 392 &  514 & $4.21_{\pm 0.05}$ & $27.5_{\pm 1.9}$ \\
RelayS2S         & Qwen2.5-7B     &  85 &   81 & $4.21_{\pm 0.05}$ & $26.9_{\pm 1.9}$ \\
\midrule
Cascaded         & Grok-4.1-fast  & 726 &  876 & $4.73_{\pm 0.03}$ & $8.4_{\pm 1.2}$  \\
RelayS2S         & Grok-4.1-fast  & 112 &   81 & $4.73_{\pm 0.03}$ & $9.4_{\pm 1.2}$  \\
\midrule
Cascaded         & GPT-4.1        & 853 & 1006 & $4.84_{\pm 0.03}$ & $5.6_{\pm 0.9}$  \\
RelayS2S         & GPT-4.1        & 119 &   81 & $4.84_{\pm 0.03}$ & $5.3_{\pm 0.9}$  \\
\midrule
\multicolumn{6}{l}{\textbf{Real voice dialogues}} \\
S2S only         & ---            &  59 &   76 & $3.17_{\pm 0.07}$ & $56.0_{\pm 2.2}$ \\
\midrule
Cascaded         & Qwen2.5-0.5B   & 205 &  239 & $3.82_{\pm 0.06}$ & $37.3_{\pm 2.1}$ \\
RelayS2S         & Qwen2.5-0.5B   & 101 &  214 & $3.81_{\pm 0.06}$ & $37.7_{\pm 2.1}$ \\
\midrule
Cascaded         & Qwen2.5-7B     & 297 &  354 & $4.06_{\pm 0.06}$ & $30.6_{\pm 2.0}$ \\
RelayS2S         & Qwen2.5-7B     & 129 &  309 & $4.03_{\pm 0.06}$ & $32.2_{\pm 2.1}$ \\
\midrule
Cascaded         & Grok-4.1-fast  & 634 &  735 & $4.54_{\pm 0.05}$ & $14.3_{\pm 1.6}$ \\
RelayS2S         & Grok-4.1-fast  & 232 &  665 & $4.52_{\pm 0.05}$ & $15.3_{\pm 1.6}$ \\
\midrule
Cascaded         & GPT-4.1        & 748 &  823 & $4.70_{\pm 0.04}$ & $9.8_{\pm 1.3}$  \\
RelayS2S         & GPT-4.1        & 269 &  655 & $4.67_{\pm 0.04}$ & $10.9_{\pm 1.4}$ \\
\bottomrule
\end{tabular}
\caption{First-chunk latency (Lat.) and response quality across system configurations on synthetic and real dialogues. RelayS2S uses a 5-word prefix throughout; the verifier commits 91.9\% and 70.7\% of prefixes on the synthetic and real-voice subsets respectively.}
\label{tab:results}
\end{table*}
 
\paragraph{Latency.}
On the synthetic subset, RelayS2S achieves a P90 first-chunk latency of
81\,ms regardless of the slow-path LLM, closely matching the 72\,ms of
the pure S2S model---the verifier adds only ${\sim}10$\,ms of overhead.
The cascaded baselines, by contrast, scale with model size, ranging from
418\,ms with Qwen2.5-0.5B up to 1{,}006\,ms with GPT-4.1.
The picture is more nuanced on real voice dialogues. The verifier
commits 70.7\% of prefixes here, so the
remaining 29.3\% of turns fall back to the cascade and pay its full
latency. P90 therefore tracks this fallback tail (655\,ms with GPT-4.1), but the bulk of committed turns run at S2S latency,
pulling the average down sharply: 748\,$\rightarrow$\,269\,ms with
GPT-4.1. The other back-ends show average latency reductions of the same character but differing magnitude --- Qwen2.5-7B (297\,$\rightarrow$\,129\,ms), and Grok-4.1-fast (634\,$\rightarrow$\,232\,ms).

\paragraph{Response quality.}
Speculative prefixing costs little quality. Across all eight configurations, RelayS2S stays within 1.6\,pp of its cascaded counterpart on low-quality rate and within 1\% on average score, while remaining far below the S2S-only baseline. The cost does not grow with back-end capacity---the real-voice gap is comparable for the 0.5B and GPT-4.1 back-ends---so a stronger slow path pays no larger price for completing a prefix it did not write. Quality is
preserved almost exactly on synthetic dialogues and degrades slightly on real voice; this tracks prefix quality rather than commit volume, since synthetic turns commit more prefixes (91.9\% vs.\ 70.7\%) yet contain far fewer bad ones (4.8\% vs.\ 17.5\%). We decompose the residual real-voice gap in \S\ref{sec:fallback_analysis}.

\paragraph{Quality--latency trade-off.}
Taken together, RelayS2S effectively decouples the quality--latency
trade-off of conventional pipelines. This pattern reveals its role as a
composition operator rather than a competitor to either
component: on synthetic dialogues, the latency gap it closes widens from
227\,ms with Qwen2.5-0.5B to 734\,ms with GPT-4.1, while committed-turn
quality tracks the cascade ceiling within $\sim$99\%. Symmetrically,
improvements to the fast path would reduce the fallback rate and tighten
the latency tail. RelayS2S thus inherits, rather than competes with,
advances on either axis, making it attractive for
deployments that rely on large or API-served LLMs where cascaded
latency would otherwise be prohibitive.
 
\paragraph{Relay margin analysis.}
\label{sec:seam_analysis}
 
While Table~\ref{tab:results} establishes that RelayS2S preserves textual response quality, the prefix-to-continuation handoff introduces an audio-level failure mode: a relay-margin stall, in which the slow path fails to deliver its first chunk before the prefix audio is exhausted. We define the relay margin as the duration of the prefix audio minus the slow-path time to first chunk. Table~\ref{tab:relay_margin} reports the distribution across different prefix lengths and LLM back-ends on committed turns.

\begin{table}[t]
\centering
\small
\setlength{\tabcolsep}{5pt}
\renewcommand{\arraystretch}{1.05}
\begin{tabular}{llrrr}
\toprule
Back-end & Prefix len & P10 (s) & Med (s) & Stall (\%) \\
\midrule
\multirow{3}{*}{Qwen2.5-7B}
& 3 words & 0.23 & 0.57 & 0.5 \\
& 5 words & 0.84 & 1.45 & 0.0 \\
& 7 words & 1.46 & 2.15 & 0.0 \\
\midrule
\multirow{3}{*}{GPT-4.1}
& 3 words & -0.19 & 0.21 & 27.5 \\
& 5 words & 0.42 & 1.03 & 1.9 \\
& 7 words & 0.96 & 1.70 & 0.5 \\
\bottomrule
\end{tabular}
\caption{Relay margin statistics on committed-prefix turns. P10 and median are computed over the per-turn relay-margin distribution.}
\label{tab:relay_margin}
\end{table}
 
Two effects stand out. The local 7B back-end is robust across all prefix lengths: even a 3-word prefix yields a positive median relay margin (0.57\,s) and a negligible stall rate (0.5\%). Conversely, API-hosted back-ends are highly sensitive to prefix length. At 3 words on GPT-4.1, the heavier latency tail pushes the lower quantile negative (P10 $=-0.19$\,s, producing audible stalls on 27.5\% of committed turns. This contextualizes why the 3-word scale used by DDTSR \cite{liu2026ddtsr} is fragile in conventional deployments: without specialized streaming mechanisms, a short prefix provides insufficient audio runway to absorb API latency tails. Extending the prefix to 5 words reduces this stall rate to 1.9\%, and 7 words eliminates stalls almost entirely.
 
\paragraph{Prefix verifier performance.}
The verifier is trained on 302K prefixes, of which 5.9\% are labeled bad.
It achieves an AUROC of 0.94 on the synthetic test set and 0.88 on the
real-speech set, where bad prefixes are considerably more common (17.5\%
vs.\ 4.8\%). At our default threshold of 0.50, the verifier commits 95.0\%
of good synthetic prefixes while rejecting 71.2\% of bad ones, giving a
fallback rate of 8.1\%. The same threshold lands at a more conservative
operating point on real speech, committing 81.1\% of good prefixes and
rejecting 78.3\% of bad ones; combined with the higher base rate, this
raises the fallback rate to 29.3\%. The full threshold sweep appears in
Appendix~\ref{app:verifier_thresholds}.
 
\paragraph{Prefix length influence.}
\label{sec:ablation_prefix}

\begin{table}[t]
\centering
\small
\renewcommand{\arraystretch}{1.15}
\setlength{\tabcolsep}{3pt}
\begin{tabular}{ccccc}
\toprule
\textbf{Prefix} & \textbf{Verifier} & \textbf{Commit} & \textbf{Quality} & \textbf{Low-Q (\%)} \\
\midrule
3 words & \ding{55} & 100\%  & $4.67_{\pm 0.04}$ & $10.9_{\pm 1.4}$ \\
3 words & \ding{51} & 87.2\% & $4.70_{\pm 0.04}$ & $ 9.8_{\pm 1.3}$ \\
\midrule
5 words & \ding{55} & 100\%  & $4.52_{\pm 0.04}$ & $14.3_{\pm 1.5}$ \\
5 words & \ding{51} & 70.7\% & $4.67_{\pm 0.04}$ & $10.9_{\pm 1.4}$ \\
\midrule
7 words & \ding{55} & 100\%  & $4.44_{\pm 0.05}$ & $16.6_{\pm 1.6}$ \\
7 words & \ding{51} & 60.8\% & $4.67_{\pm 0.04}$ & $10.5_{\pm 1.4}$ \\
\bottomrule
\end{tabular}
\caption{Ablation on prefix length and verifier gating on real voice subset with the GPT-4.1 back-end.}
\label{tab:ablation}
\end{table}

Table~\ref{tab:ablation} ablates prefix length against verifier gating. Without gating, quality degrades monotonically as prefixes lengthen: more S2S content is committed before the stronger LLM can intervene, and the low-quality rate rises by 5.7\,pp from 3 to 7 words. This isolates the cost of un-gated early commitment, the regime KAME operates in \cite{kuroki2025kame}. With the verifier enabled, that cost is contained: the low-quality rate stays within a 1.1\,pp band across all prefix lengths, and the protection widens as
prefixes grow (1.1\,pp at 3 words, 3.4 at 5, 6.1 at 7). Gating is what turns prefix length from a quality dial into a free parameter. Quality alone would favour the shortest prefix---gated 3-word attains the best low-quality rate at the highest commit rate---but three words leave too little audio runway to absorb the GPT-4.1 latency tail (27.5\% stalls, Table~\ref{tab:relay_margin}).
 
\paragraph{Fallback decomposition.}
\label{sec:fallback_analysis}
 
\begin{table}[t]
\centering
\small
\setlength{\tabcolsep}{4pt}
\begin{tabular*}{\columnwidth}{lr@{\extracolsep{\fill}}cccc}
\toprule
 & & \multicolumn{2}{c}{\textbf{Avg Quality}} & \multicolumn{2}{c}{\textbf{Low-Q (\%)}} \\
\cmidrule(lr){3-4} \cmidrule(lr){5-6}
\textbf{Subset} & \multicolumn{1}{c}{$n$} & \textbf{Casc.} & \textbf{RelayS2S} & \textbf{Casc.} & \textbf{RelayS2S} \\
\midrule
Committed & 1,414 & 4.83 & 4.79 & 5.5  & 7.1  \\
Fallback  &   586 & 4.36 & 4.38 & 20.3 & 20.1 \\
\midrule
All       & 2,000 & 4.70 & 4.67 & 9.8  & 10.9 \\
\bottomrule
\end{tabular*}
\caption{Fallback quality decomposition with GPT-4.1 as the back-end on real speech ($n{=}2{,}000$).}
\label{tab:fallback-decomposition}
\end{table}
 
Table~\ref{tab:fallback-decomposition} confirms that the verifier correctly segregates turns by macro-level difficulty: the cascaded pipeline scores significantly lower on the subset of turns triggered for fallback than on committed turns (4.36 vs.\ 4.83) and yields a nearly $4\times$ higher low-quality rate (20.3\% vs.\ 5.5\%). Fallback correlates with acoustic difficulty: ASR WER on fallback turns is
roughly three times that on committed turns (8.7\% vs.\ 2.8\%). But it is
not the whole story---over half of fallback turns are transcribed perfectly,
and these are largely cases where the fast path mishandles rare names,
entities, or technical terms. This explains our latency profile: RelayS2S captures S2S-latency speed on the 70.7\% of easier, committed turns while gracefully absorbing full cascaded latency on the harder 29.3\% of fallbacks. To isolate whether the residual 1.1\% aggregate quality gap stems from prefix propagation or baseline difficulty, we cross-tabulate human prefix-sensibility labels against final response quality on committed turns (Appendix~\ref{app:contingency}). Only 5.4\% of committed turns contain a non-sensible prefix; crucially, on these difficult turns, the slow path actively recovers to generate an acceptable response in 63.2\% of cases. This demonstrates that prefix-conditioned generation actively repairs, rather than passively propagates, flawed prefixes. The remaining 1.98\% of committed turns where both the prefix and response fail represents the irreducible cost of selective handoff, accounting for the bulk of RelayS2S's committed-turn delta.

\section{Conclusion}

We presented RelayS2S, a speculative hybrid architecture that pairs a
duplex S2S model with a cascaded ASR--LLM pipeline to break the
latency--quality trade-off in real-time spoken dialogue. The fast path
drafts a short response prefix via forked speculative generation,
decoupling rapid drafting from live audio monitoring. A lightweight learned verifier then
gates the handoff, committing the prefix when it is contextually safe
and falling back to the cascade otherwise.

Across four slow-path back-ends, RelayS2S approaches the latency profile
of a pure S2S model while retaining ${\sim}99\%$ of the cascaded
pipeline's textual quality. With GPT-4.1, P90 first-chunk latency drops
from 1{,}006\,ms to 81\,ms on synthetic dialogues, and average
first-chunk latency drops by 479\,ms on real voice
dialogues. More fundamentally, this pattern reveals RelayS2S's role as a composition operator over its components: advances on either axis propagate directly into the system, narrowing the latency–quality gap between S2S and cascaded endpoints without architectural change.

\section*{Limitations}

RelayS2S has several limitations worth noting. First, the latency tail remains
dominated by fallback events: 29.3\% of real-voice turns are routed through the
full cascade, so P90 latency with a GPT-4.1 back-end is 655\,ms, against 81\,ms
on synthetic data. Fallback is triggered mainly by poor prefix drafting from the
fast-path S2S model, and training it on more diverse acoustic and textual data
would likely tighten the tail. Second, the system is vulnerable to relay-margin
stalls, where the slow path fails to deliver its first continuation chunk before
the prefix audio ends. This is worst with API-hosted LLMs: a 3-word prefix with
GPT-4.1 stalls on 27.5\% of committed turns, which extending to 7 words
mitigates. Filler speech or adaptive prefix length could absorb the remainder.
Third, our evaluation is heavily automated. We run no user study of perceived
responsiveness, prosodic continuity at the handoff, or interruption naturalness.
Our real-voice test set also pairs authentic user audio with synthesized
assistant turns rather than naturally occurring dialogue. Fourth, all data are
English. The 5-word prefix was chosen because it yields roughly two seconds of
English speech, so the relay-margin budget would need to be re-derived for
languages with different information density or without orthographic word
boundaries.

\section*{Ethical Considerations}

\textbf{Data compliance and asset use.}
All speech corpora used in this work---GigaSpeech, Common Voice,
Switchboard, and TAU Urban Acoustic Scenes---are publicly
released for research use, and we use them strictly within the terms of their licenses. We do not redistribute raw audio data. 

\noindent \textbf{Deployment risks and mitigation.}
By design, RelayS2S initiates speech generation before the slow path has reasoned over the complete user utterance. Although the verifier successfully rejects most invalid prefixes and the slow path dynamically repairs committed errors (Appendix~\ref{app:contingency}), a small residual fraction of turns may still produce an audible error before a correction can be executed. Consequently, we caution against deploying this system in safety-sensitive domains (e.g., medical, legal, or mental health applications) without safeguards, such as domain-specific verifiers, more conservative operating thresholds, or restricting speculative generation to non-substantive response openers.

\noindent \textbf{Scientific use of AI assistants.} 
Large language models were utilized during the development of this project to assist with code optimization, conceptual brainstorming, and stylistic polishing of the manuscript. The core research directions, experimental designs, and data interpretations were conceived and directed entirely by the human authors, who maintain full accountability for the scientific integrity and accuracy of the work.

\bibliography{custom}

@article{levinson2015timing,
    
AUTHOR={Levinson, Stephen C.  and Torreira, Francisco },
           
TITLE={Timing in turn-taking and its implications for processing models of language},
          
JOURNAL={Frontiers in Psychology},
          
VOLUME={Volume 6 - 2015},
  
YEAR={2015},

DOI={10.3389/fpsyg.2015.00731},
  
ISSN={1664-1078}}

@inproceedings{zhang2023speechgpt,
  title={Speechgpt: Empowering large language models with intrinsic cross-modal conversational abilities},
  author={Zhang, Dong and Li, Shimin and Zhang, Xin and Zhan, Jun and Wang, Pengyu and Zhou, Yaqian and Qiu, Xipeng},
  booktitle={Findings of the Association for Computational Linguistics: EMNLP 2023},
  pages={15757--15773},
  year={2023}
}

@misc{rubenstein2023audiopalm,
      title={AudioPaLM: A Large Language Model That Can Speak and Listen}, 
      author={Paul K. Rubenstein and Chulayuth Asawaroengchai and Duc Dung Nguyen and Ankur Bapna and Zalán Borsos and Félix de Chaumont Quitry and Peter Chen and Dalia El Badawy and Wei Han and Eugene Kharitonov and Hannah Muckenhirn and Dirk Padfield and James Qin and Danny Rozenberg and Tara Sainath and Johan Schalkwyk and Matt Sharifi and Michelle Tadmor Ramanovich and Marco Tagliasacchi and Alexandru Tudor and Mihajlo Velimirović and Damien Vincent and Jiahui Yu and Yongqiang Wang and Vicky Zayats and Neil Zeghidour and Yu Zhang and Zhishuai Zhang and Lukas Zilka and Christian Frank},
      year={2023},
      eprint={2306.12925},
      archivePrefix={arXiv},
      primaryClass={cs.CL},
      url={https://arxiv.org/abs/2306.12925}, 
}

@misc{defossez2024moshi,
      title={Moshi: a speech-text foundation model for real-time dialogue}, 
      author={Alexandre Défossez and Laurent Mazaré and Manu Orsini and Amélie Royer and Patrick Pérez and Hervé Jégou and Edouard Grave and Neil Zeghidour},
      year={2024},
      eprint={2410.00037},
      archivePrefix={arXiv},
      primaryClass={eess.AS},
      url={https://arxiv.org/abs/2410.00037}, 
}

@misc{nguyen2025salm,
      title={SALM-Duplex: Efficient and Direct Duplex Modeling for Speech-to-Speech Language Model}, 
      author={Ke Hu and Ehsan Hosseini-Asl and Chen Chen and Edresson Casanova and Subhankar Ghosh and Piotr Żelasko and Zhehuai Chen and Jason Li and Jagadeesh Balam and Boris Ginsburg},
      year={2025},
      eprint={2505.15670},
      archivePrefix={arXiv},
      primaryClass={cs.CL},
      url={https://arxiv.org/abs/2505.15670}, 
}

@misc{li2025personaplex,
      title={PersonaPlex: Voice and Role Control for Full Duplex Conversational Speech Models}, 
      author={Rajarshi Roy and Jonathan Raiman and Sang-gil Lee and Teodor-Dumitru Ene and Robert Kirby and Sungwon Kim and Jaehyeon Kim and Bryan Catanzaro},
      year={2026},
      eprint={2602.06053},
      archivePrefix={arXiv},
      primaryClass={cs.CL},
      url={https://arxiv.org/abs/2602.06053}, 
}

@misc{chen2025fireredchat,
      title={FireRedChat: A Pluggable, Full-Duplex Voice Interaction System with Cascaded and Semi-Cascaded Implementations}, 
      author={Junjie Chen and Yao Hu and Junjie Li and Kangyue Li and Kun Liu and Wenpeng Li and Xu Li and Ziyuan Li and Feiyu Shen and Xu Tang and Manzhen Wei and Yichen Wu and Fenglong Xie and Kaituo Xu and Kun Xie},
      year={2025},
      eprint={2509.06502},
      archivePrefix={arXiv},
      primaryClass={cs.SD},
      url={https://arxiv.org/abs/2509.06502}, 
}

@misc{liu2025xtalk,
      title={X-Talk: On the Underestimated Potential of Modular Speech-to-Speech Dialogue System}, 
      author={Zhanxun Liu and Yifan Duan and Mengmeng Wang and Pengchao Feng and Haotian Zhang and Xiaoyu Xing and Yijia Shan and Haina Zhu and Yuhang Dai and Chaochao Lu and Xipeng Qiu and Lei Xie and Lan Wang and Nan Yan and Zilong Zheng and Ziyang Ma and Kai Yu and Xie Chen},
      year={2025},
      eprint={2512.18706},
      archivePrefix={arXiv},
      primaryClass={cs.SD},
      url={https://arxiv.org/abs/2512.18706}, 
}

@misc{zou2026lts,
      title={LTS-VoiceAgent: A Listen-Think-Speak Framework for Efficient Streaming Voice Interaction via Semantic Triggering and Incremental Reasoning}, 
      author={Wenhao Zou and Yuwei Miao and Zhanyu Ma and Jun Xu and Jiuchong Gao and Jinghua Hao and Renqing He and Jingwen Xu},
      year={2026},
      eprint={2601.19952},
      archivePrefix={arXiv},
      primaryClass={cs.SD},
      url={https://arxiv.org/abs/2601.19952}, 
}

@misc{liu2026ddtsr,
      title={Discourse-Aware Dual-Track Streaming Response for Low-Latency Spoken Dialogue Systems}, 
      author={Siyuan Liu and Jiahui Xu and Feng Jiang and Kuang Wang and Zefeng Zhao and Chu-Ren Huang and Jinghang Gu and Changqing Yin and Haizhou Li},
      year={2026},
      eprint={2602.23266},
      archivePrefix={arXiv},
      primaryClass={cs.CL},
      url={https://arxiv.org/abs/2602.23266}, 
}

@misc{kuroki2025kame,
      title={KAME: Tandem Architecture for Enhancing Knowledge in Real-Time Speech-to-Speech Conversational AI}, 
      author={So Kuroki and Yotaro Kubo and Takuya Akiba and Yujin Tang},
      year={2026},
      eprint={2510.02327},
      archivePrefix={arXiv},
      primaryClass={cs.CL},
      url={https://arxiv.org/abs/2510.02327}, 
}

@misc{likhomanenko2025chipchat,
      title={ChipChat: Low-Latency Cascaded Conversational Agent in MLX}, 
      author={Tatiana Likhomanenko and Richard He Bai and Zijin Gu and Zakaria Aldeneh and Shiladitya Dutta and Luke Carlson and Han Tran and Yizhe Zhang and Ruixiang Zhang and Huangjie Zheng and Navdeep Jaitly},
      year={2026},
      eprint={2509.00078},
      archivePrefix={arXiv},
      primaryClass={eess.AS},
      url={https://arxiv.org/abs/2509.00078}, 
}

@misc{voiceassistant,
      title={Mini-Omni: Language Models Can Hear, Talk While Thinking in Streaming}, 
      author={Zhifei Xie and Changqiao Wu},
      year={2024},
      eprint={2408.16725},
      archivePrefix={arXiv},
      primaryClass={cs.AI},
      url={https://arxiv.org/abs/2408.16725}, 
}

@article{openmoss,
  title   = {{MOSS}: An Open Conversational Large Language Model},
  author  = {Sun, Tianxiang and Zhang, Xiaotian and He, Zhengfu and Li, Peng
             and Cheng, Qinyuan and Liu, Xiangyang and Yan, Hang and Shao, Yunfan
             and Tang, Qiong and Zhang, Shiduo and Zhao, Xingjian and Chen, Ke
             and Zheng, Yining and Zhou, Zhejian and Li, Ruixiao and Zhan, Jun
             and Zhou, Yunhua and Li, Linyang and Yang, Xiaogui and Wu, Lingling
             and Yin, Zhangyue and Huang, Xuanjing and Jiang, Yu-Gang and Qiu, Xipeng},
  journal = {Machine Intelligence Research},
  year    = {2024},
  month   = oct,
  doi     = {10.1007/s11633-024-1502-8},
  url     = {https://doi.org/10.1007/s11633-024-1502-8},
  publisher = {Springer}
}

@inproceedings{topical,
  title     = {{Topical-Chat: Towards Knowledge-Grounded Open-Domain Conversations}},
  author    = {Karthik Gopalakrishnan and Behnam Hedayatnia and Qinlang Chen and Anna Gottardi and Sanjeev Kwatra and Anu Venkatesh and Raefer Gabriel and Dilek Hakkani-Tür},
  year      = {2019},
  booktitle = {{Interspeech 2019}},
  pages     = {1891--1895},
  doi       = {10.21437/Interspeech.2019-3079},
  issn      = {2958-1796},
}

@misc{convai2,
      title={The Second Conversational Intelligence Challenge (ConvAI2)}, 
      author={Emily Dinan and Varvara Logacheva and Valentin Malykh and Alexander Miller and Kurt Shuster and Jack Urbanek and Douwe Kiela and Arthur Szlam and Iulian Serban and Ryan Lowe and Shrimai Prabhumoye and Alan W Black and Alexander Rudnicky and Jason Williams and Joelle Pineau and Mikhail Burtsev and Jason Weston},
      year={2019},
      eprint={1902.00098},
      archivePrefix={arXiv},
      primaryClass={cs.AI},
      url={https://arxiv.org/abs/1902.00098}, 
}

@inproceedings{bst,
    title = "Can You Put it All Together: Evaluating Conversational Agents' Ability to Blend Skills",
    author = "Smith, Eric Michael  and
      Williamson, Mary  and
      Shuster, Kurt  and
      Weston, Jason  and
      Boureau, Y-Lan",
    booktitle = "Proceedings of the 58th Annual Meeting of the Association for Computational Linguistics",
    month = jul,
    year = "2020",
    address = "Online",
    publisher = "Association for Computational Linguistics",
    url = "https://aclanthology.org/2020.acl-main.183/",
    doi = "10.18653/v1/2020.acl-main.183",
    pages = "2021--2030",
}

@misc{cosyvoice,
      title={CosyVoice 2: Scalable Streaming Speech Synthesis with Large Language Models}, 
      author={Zhihao Du and Yuxuan Wang and Qian Chen and Xian Shi and Xiang Lv and Tianyu Zhao and Zhifu Gao and Yexin Yang and Changfeng Gao and Hui Wang and Fan Yu and Huadai Liu and Zhengyan Sheng and Yue Gu and Chong Deng and Wen Wang and Shiliang Zhang and Zhijie Yan and Jingren Zhou},
      year={2024},
      eprint={2412.10117},
      archivePrefix={arXiv},
      primaryClass={cs.SD},
      url={https://arxiv.org/abs/2412.10117}, 
}

@misc{tau_noise,
  author       = {Heittola, Toni Kristian and
                  Mesaros, Annamaria and
                  Virtanen, Tuomas},
  title        = {TAU Urban Acoustic Scenes 2024 Mobile, Evaluation dataset },
  month        = jun,
  year         = 2024,
  publisher    = {Zenodo},
  doi          = {10.5281/zenodo.11366913},
  url          = {https://doi.org/10.5281/zenodo.11366913},
}

@misc{qwen2.5,
      title={Qwen2.5 Technical Report}, 
      author={{Qwen Team} and An Yang and Baosong Yang and Beichen Zhang and Binyuan Hui and Bo Zheng and Bowen Yu and Chengyuan Li and Dayiheng Liu and Fei Huang and Haoran Wei and Huan Lin and Jian Yang and Jianhong Tu and Jianwei Zhang and Jianxin Yang and Jiaxi Yang and Jingren Zhou and Junyang Lin and Kai Dang and Keming Lu and Keqin Bao and Kexin Yang and Le Yu and Mei Li and Mingfeng Xue and Pei Zhang and Qin Zhu and Rui Men and Runji Lin and Tianhao Li and Tianyi Tang and Tingyu Xia and Xingzhang Ren and Xuancheng Ren and Yang Fan and Yang Su and Yichang Zhang and Yu Wan and Yuqiong Liu and Zeyu Cui and Zhenru Zhang and Zihan Qiu},
      year={2025},
      eprint={2412.15115},
      archivePrefix={arXiv},
      primaryClass={cs.CL},
      url={https://arxiv.org/abs/2412.15115}, 
}

@misc{faster_whisper,
      title={Robust Speech Recognition via Large-Scale Weak Supervision}, 
      author={Alec Radford and Jong Wook Kim and Tao Xu and Greg Brockman and Christine McLeavey and Ilya Sutskever},
      year={2022},
      eprint={2212.04356},
      archivePrefix={arXiv},
      primaryClass={eess.AS},
      url={https://arxiv.org/abs/2212.04356}, 
}

@InProceedings{forzen_llm,
  title = 	 {Freeze-Omni: A Smart and Low Latency Speech-to-speech Dialogue Model with Frozen {LLM}},
  author =       {Wang, Xiong and Li, Yangze and Fu, Chaoyou and Zhang, Yike and Shen, Yunhang and Xie, Lei and Li, Ke and Sun, Xing and Ma, Long},
  booktitle = 	 {Proceedings of the 42nd International Conference on Machine Learning},
  pages = 	 {63345--63354},
  year = 	 {2025},
  editor = 	 {Singh, Aarti and Fazel, Maryam and Hsu, Daniel and Lacoste-Julien, Simon and Berkenkamp, Felix and Maharaj, Tegan and Wagstaff, Kiri and Zhu, Jerry},
  volume = 	 {267},
  series = 	 {Proceedings of Machine Learning Research},
  month = 	 {13--19 Jul},
  publisher =    {PMLR},
  url = 	 {https://proceedings.mlr.press/v267/wang25aw.html},
}

@misc{speakstream2025,
      title={SpeakStream: Streaming Text-to-Speech with Interleaved Data}, 
      author={Richard He Bai and Zijin Gu and Tatiana Likhomanenko and Navdeep Jaitly},
      year={2025},
      eprint={2505.19206},
      archivePrefix={arXiv},
      primaryClass={cs.CL},
      url={https://arxiv.org/abs/2505.19206}, 
}

@inproceedings{gigaspeech,
  title     = {{GigaSpeech: An Evolving, Multi-Domain ASR Corpus with 10,000 Hours of Transcribed Audio}},
  author    = {Guoguo Chen and Shuzhou Chai and Guan-Bo Wang and Jiayu Du and Wei-Qiang Zhang and Chao Weng and Dan Su and Daniel Povey and Jan Trmal and Junbo Zhang and Mingjie Jin and Sanjeev Khudanpur and Shinji Watanabe and Shuaijiang Zhao and Wei Zou and Xiangang Li and Xuchen Yao and Yongqing Wang and Zhao You and Zhiyong Yan},
  year      = {2021},
  booktitle = {{Interspeech 2021}},
  pages     = {3670--3674},
  doi       = {10.21437/Interspeech.2021-1965},
  issn      = {2958-1796},
}

@inproceedings{common,
  title={Common voice: A massively-multilingual speech corpus},
  author={Ardila, Rosana and Branson, Megan and Davis, Kelly and Kohler, Michael and Meyer, Josh and Henretty, Michael and Morais, Reuben and Saunders, Lindsay and Tyers, Francis and Weber, Gregor},
  booktitle={Proceedings of the twelfth language resources and evaluation conference},
  pages={4218--4222},
  year={2020}
}

@INPROCEEDINGS{switchboard,
  author={Godfrey, J.J. and Holliman, E.C. and McDaniel, J.},
  booktitle={[Proceedings] ICASSP-92: 1992 IEEE International Conference on Acoustics, Speech, and Signal Processing}, 
  title={SWITCHBOARD: telephone speech corpus for research and development}, 
  year={1992},
  volume={1},
  number={},
  pages={517-520 vol.1},
  doi={10.1109/ICASSP.1992.225858}}

@misc{gpt5,
  author       = {{OpenAI}},
  title        = {Introducing {GPT-5.4}},
  year         = {2026},
  month        = mar,
  howpublished = {\url{https://openai.com/index/introducing-gpt-5-4/}},
}

@misc{gpt41,
  author       = {{OpenAI}},
  title        = {Introducing {GPT-4.1} in the {API}},
  year         = {2025},
  month        = apr,
  howpublished = {\url{https://openai.com/index/gpt-4-1/}},
}

@misc{grok41,
  author       = {{xAI}},
  title        = {Grok 4.1},
  year         = {2025},
  month        = nov,
  howpublished = {\url{https://x.ai/news/grok-4-1}},
}

@misc{elevenlabs_flash_v25,
  author       = {{ElevenLabs}},
  title        = {Models: {Flash} v2.5},
  year         = {2025},
  howpublished = {\url{https://elevenlabs.io/docs/overview/models}},
}

@misc{gemini3flash,
  author       = {{Google}},
  title        = {Gemini 3 Flash},
  year         = {2025},
  month        = dec,
  howpublished = {\url{https://blog.google/products-and-platforms/products/gemini/gemini-3-flash/}},
}

@misc{gemini31pro,
  author       = {{Google}},
  title        = {Gemini 3.1 {Pro}},
  year         = {2026},
  howpublished = {\url{https://deepmind.google/models/gemini/pro/}},
}

@misc{pred_gen,
      title={PredGen: Accelerated Inference of Large Language Models through Input-Time Speculation for Real-Time Speech Interaction}, 
      author={Shufan Li and Aditya Grover},
      year={2025},
      eprint={2506.15556},
      archivePrefix={arXiv},
      primaryClass={cs.CL},
      url={https://arxiv.org/abs/2506.15556}, 
}

@misc{gemma4model,
  author       = {{Google}},
  title        = {Gemma 4},
  year         = {2026},
  howpublished = {\url{https://deepmind.google/models/gemma/gemma-4/}},
}

@misc{qwen36model,
  author       = {{Qwen Team}},
  title        = {Qwen3.6},
  year         = {2026},
  howpublished = {\url{https://qwen.ai/blog?id=qwen3.6-27b}},
}

@misc{gptoss120b,
  author       = {{OpenAI}},
  title        = {Introducing gpt-oss},
  year         = {2025},
  howpublished = {\url{https://openai.com/index/introducing-gpt-oss/}},
}

\appendix
 
\section{Appendix}

\subsection{Released-Checkpoint Evaluation of KAME}
\label{app:kame}

We evaluate the publicly released KAME checkpoint in the closest setting
available to our experiments: a 7B Moshi S2S front-end with a GPT-4.1
cascaded back-end, on 1{,}000 single-turn synthetic speech prompts drawn
from our test set. Table~\ref{tab:kame} reports response quality under the
same rubric used throughout (\S\ref{sec:metrics}).

\begin{table}[h]
\centering
\small
\begin{tabular}{lcc}
\toprule
\textbf{System} & \textbf{Avg Quality} & \textbf{Low-quality (\%)} \\
\midrule
KAME     & 2.49 & 71.1 \\
RelayS2S & 4.81 &  5.7 \\
\bottomrule
\end{tabular}
\caption{Released-checkpoint evaluation of KAME against RelayS2S on 1{,}000
single-turn synthetic speech prompts (GPT-4.1 back-end).}
\label{tab:kame}
\end{table}

This is not a controlled architectural comparison and should not be read as
evidence that the KAME architecture is intrinsically inferior. The released
checkpoint is trained on different data and targets limited reasoning tasks
such as MMLU-Pro and GSM8K, whereas our test set consists of open-ended
conversational prompts; performance may also depend on interface and
prompting choices.

To diagnose the gap, we inspected KAME's intermediate outputs. On the 578
turns where both the ASR transcript and the back-end response guidance are
judged clean, the final KAME output still scores 2.74/5 on average. This
suggests the gap stems primarily from KAME's use of the front-end S2S model
as the primary response generator, with the cascaded back-end
providing only soft guidance: the front-end is susceptible to distribution
shift, producing word-level corruption, weak adherence to the
back-end-generated response, post-answer rambling, and occasional silent
outputs. RelayS2S instead makes the back-end
authoritative for everything after the committed prefix, which bounds how
much front-end error can reach the listener.

\subsection{Training Data Details}
\label{app:training_data}

We construct a hybrid training corpus from two sources: (1)~fully
synthetic dialogues rendered entirely with text-to-speech (TTS) systems,
and (2)~real voice dialogues combining authentic human speech recordings
with synthetic assistant responses.

\subsection{Synthetic Voice Dialogues}
\label{app:data_synthetic}

We source foundational text dialogues from two channels. First, we collect single- and multi-turn user--assistant conversations from VoiceAssistant~\cite{voiceassistant} and OpenMOSS~\cite{openmoss}. Second, we prompt GPT-5.4 to generate additional conversations using topic and plot outlines derived from human--human dialogue corpora, including TopicalChat~\cite{topical}, ConvAI~\cite{convai2}, and BlendedSkillTalk~\cite{bst}. All dialogues are synthesized into 16\,kHz speech using CosyVoice2~\cite{cosyvoice} with a default inter-turn silence of 200\,ms.

To improve robustness to real-world acoustic conditions, we mix in non-overlapping noise segments sampled from the TAU Urban Acoustic Scenes dataset~\cite{tau_noise}. Noise clips span 5--10\,s, cover 50--100\% of each conversation, and are applied at signal-to-noise ratios uniformly sampled between 0 and 20\,dB.

The resulting synthetic corpus contains 104K conversations totaling 2{,}133 hours of audio, with conversation lengths ranging from 3\,s to 3\,mins.

\subsection{Real Voice Dialogues}
\label{app:data_semi_synthetic}

To complement the synthetic data with authentic vocal acoustics, speaking styles, and recording conditions, we incorporate real human speech from three large-scale automatic speech recognition (ASR) corpora: GigaSpeech~\cite{gigaspeech}, Common Voice~\cite{common}, and Switchboard~\cite{switchboard}.

Because raw ASR transcripts often contain incomplete conversational fragments, narrative speech, or context-dependent utterances unsuitable for assistant interactions, we apply an automated filtering pipeline. Specifically, we use Gemini-3-Flash to classify each transcript for whether it is both context-independent and plausibly directed at an AI assistant (see Appendix~\ref{app:real_voice} for prompts and filtering criteria).

This process yields approximately 90{,}000 curated human audio prompts. We then prompt GPT-5.4 to generate instruction-following assistant responses conditioned on their transcripts, yielding 90K two-turn real-voice conversations totaling 304 hours of audio. The synthetic assistant responses here are not treated as gold references; system outputs are evaluated reference-free against the dialogue context.

\subsection{Duplex Phenomena Injection}
\label{app:data_duplex}

To train the duplex control tokens described in \S\ref{sec:fast_path}, we augment the synthetic dialogues with three categories of overlapping-speech events.

\paragraph{Backchannels.}
We prompt Gemini-3.1-Pro \cite{gemini31pro} to identify appropriate insertion points and generate backchannel utterances (e.g., \textit{uh-huh}, \textit{right}) during user speech.

\begin{figure}[t]
    \centering
    \includegraphics[width=\columnwidth]{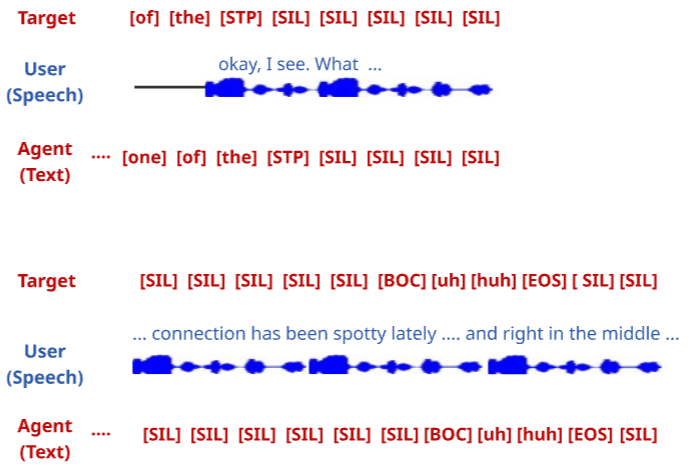}
    \caption{Training examples for duplex control tokens.
    Top: the agent emits \texttt{[STP]} when interrupted by the user.
    Bottom: the agent emits \texttt{[BOC]} to produce a backchannel during user speech.}
    \label{fig:duplex_examples}
\end{figure}

\paragraph{Interruptions.}
We randomly truncate assistant utterances between 20\% and 60\% of their duration, then overlap the following user turn by 320\,ms to simulate barge-in behavior. Figure~\ref{fig:duplex_examples} shows examples of interruption and backchannel events.

\paragraph{Mid-utterance pauses.}
We insert \texttt{[PAUSE]} tokens into user utterances to simulate extended hesitations, exposing the model to natural pause patterns and discouraging premature responses.

\subsection{Implementation Details}
\label{app:implementation}

\paragraph{Fast-path S2S model.}
The speech encoder is a 24-block conformer with 1024-dimensional hidden states,
16 attention heads, and relative position encodings. It consumes
80-dimensional log-Mel features extracted at 16\,kHz with a 10\,ms frame shift
and applies $4\times$ subsampling internally. The encoder operates with a chunk
size of 4 subsampled frames and 16 left-context chunks. The CNN adapter consists
of two causal 1D convolutional stages (kernel size~5, stride~2, left-padded).
The first stage preserves the 1024-dimensional encoder output; the second
projects down to 896 dimensions to match the LLM hidden size. Together with the
encoder's internal $4\times$ subsampling, the two stride-2 adapter stages yield
an overall $16\times$ temporal downsampling from $10\,\mathrm{ms}$ input frames,
producing one speech representation every $160\,\mathrm{ms}$. The LLM backbone
is Qwen2.5-0.5B~\cite{qwen2.5}.

\paragraph{Training procedure.}
Training proceeds in three stages:
(1)~\textit{Encoder--adapter pretraining}: the LLM backbone is frozen and only
the speech encoder and CNN adapter are updated, using speech transcription as
the learning objective so that the adapted representations align with the
LLM's text embedding space.
(2)~\textit{Full-duplex S2S training}: all parameters are unfrozen and the
model is trained end-to-end on the synthetic and real voice conversations data described
in \S\ref{sec:data}.
(3)~\textit{Prefix verifier training}, following the procedure described in
\S\ref{sec:gating}.
All three stages are run on two NVIDIA L40S GPUs (48\,GB VRAM each), with each
stage converging within 48 hours.

\subsection{Verifier Architecture}
\label{app:verifier_arch}

This section provides the full architectural specification of the prefix
verifier described in \S\ref{sec:gating}. Given the per-position decoder
hidden state $\mathbf{h}_t \in \mathbb{R}^{d}$ and scalar feature vector
$\mathbf{s}_t \in \mathbb{R}^{3}$, the hidden state at each position is
first projected to a lower-dimensional space:
\begin{equation}
    \hat{\mathbf{h}}_t
    = \mathrm{GELU}(\mathrm{LayerNorm}(W_p\,\mathbf{h}_t + b_p))
\end{equation}
where $W_p \in \mathbb{R}^{d' \times d}$ with $d' \ll d$. The scalar features
then modulate this representation through a \textit{vector gate}:
\begin{equation}
    \tilde{\mathbf{z}}_t
    = \hat{\mathbf{h}}_t \odot \sigma(W_g\,\mathbf{s}_t + b_g)
      \;+\; W_a\,\mathbf{s}_t + b_a
\end{equation}
where $W_g, W_a \in \mathbb{R}^{d' \times 3}$. The multiplicative gate
controls how much each hidden dimension is trusted given the calibration
signals, while the additive bypass ensures scalar information flows even when
the gate saturates. Learned positional embeddings are added, followed by a
token-level feedforward block with a residual connection:
\begin{equation}
    \mathbf{z}_t = \tilde{\mathbf{z}}_t + \mathrm{pos}_t + \mathrm{FF}(\tilde{\mathbf{z}}_t + \mathrm{pos}_t)
\end{equation}
The sequence $\mathbf{z}_{1:n}$ is aggregated via attention pooling with a
single learned query $\mathbf{q}$ and learned key/value projections
$W_k, W_v \in \mathbb{R}^{d' \times d'}$:
\begin{equation}
    \mathbf{p}
    = \mathrm{LayerNorm}\!\bigl(\mathrm{Attn}(\mathbf{q},\,
      W_k\,\mathbf{z}_{1:n},\,W_v\,\mathbf{z}_{1:n})\bigr)
\end{equation}
and passed through a linear head followed by a sigmoid to produce a confidence
score $c \in [0,1]$.

\subsection{Judge Validation Details}
\label{app:judge_validation}

This section details the validation of the Gemini-3-Flash judge used to score textual response quality (\S\ref{sec:metrics}). The judge is given the reference dialogue transcript, not the system ASR transcript, so ASR-induced misunderstandings are reflected in the response-quality score.

\paragraph{Run-to-run stability.}
We analyzed the judge's variance across three independent runs on a
1{,}000-sample evaluation set. The aggregate low-quality rate varied by
only 0.47 percentage points (standard deviation) across runs, yielding
high inter-run consistency (Fleiss' $\kappa = 0.81$ for the binary
label), which is comparable to the run-to-run variation of the average
quality score ($\sigma = 0.016$). While 13.9\% of individual items
received inconsistent binary classifications across evaluations (mean
per-item $\sigma = 0.21$), these label flips were symmetric and
neutralized at the population level.

\paragraph{Validation against human annotators.}
To validate the Gemini-3-Flash judge against human standards, we
compared its binarized low-quality decisions on the same 1{,}000-sample
set against the majority vote of a panel  consisting of three human annotators. The
automated judge achieved a strong Cohen's $\kappa = 0.70$ against the
human consensus, with a precision of 0.96 and a recall of 0.83 for the
low-quality class. Within the panel, individual annotators agreed with
the majority at $\kappa \geq 0.78$, confirming high internal alignment. The panel was 
comprised of three expert human annotators with backgrounds in computer science, who participated on a voluntary basis. The evaluation 
protocol did not involve the collection or storage of any personal data; 
annotators were exclusively tasked with providing objective classifications 
of model output quality, ensuring no personally identifiable information 
or sensitive participant data was captured.

\subsection{Prefix Verifier Threshold Sweep}
\label{app:verifier_thresholds}

Table~\ref{tab:verifier_thresholds} shows the tradeoff between rejecting
bad prefixes and preserving fast-path coverage across operating
thresholds. Lower thresholds maximize latency savings by committing
nearly all prefixes, while higher thresholds reduce bad commits at the
cost of more frequent fallback to the slower cascaded pipeline. We select a threshold of 0.50 as the default operating point. On
synthetic data, this threshold rejects 71\% of bad prefixes while
committing 95\% of good prefixes, with an overall fallback rate of 8\%.
On real speech, it rejects 78\% of bad prefixes while committing 81\% of
good prefixes, with a 29\% fallback rate. Increasing the threshold to
0.75 nearly eliminates bad commits (1--2\%) but substantially reduces
fast-path coverage, especially on real speech.

\begin{table}[t]
\centering
\scriptsize
\setlength{\tabcolsep}{4pt}
\renewcommand{\arraystretch}{1.1}

\resizebox{\columnwidth}{!}{
\begin{tabular}{lcccc}
\toprule
\textbf{Subset} & \textbf{Thr.} & \textbf{Bad reject} & \textbf{Good commit} & \textbf{Fallback} \\
\midrule
\multirow{3}{*}{Synthetic}
 & 0.25 & 24.0\% & 99.7\% &  1.4\% \\
 & 0.50 & 71.2\% & 95.0\% &  8.1\% \\
 & 0.75 & 98.1\% & 70.1\% & 33.2\% \\
\midrule
\multirow{3}{*}{Real}
 & 0.25 & 30.9\% & 98.8\% &  6.4\% \\
 & 0.50 & 78.3\% & 81.1\% & 29.3\% \\
 & 0.75 & 98.6\% & 27.6\% & 76.9\% \\
\bottomrule
\end{tabular}
}

\caption{Prefix verifier operating points on synthetic and real-speech test sets.}
\label{tab:verifier_thresholds}
\end{table}

\subsection{Prefix-to-Response Contingency Analysis}
\label{app:contingency}

To isolate whether RelayS2S's residual 1.1-percentage-point gap on committed
turns (10.9\% vs.\ 9.8\% low-quality, Table~\ref{tab:fallback-decomposition})
stems from prefix propagation errors or baseline turn difficulty, we
cross-tabulate human prefix-sensibility labels against final response
quality within the committed subset. Table~\ref{tab:contingency} reports
the resulting contingency matrix on real speech with the GPT-4.1
back-end.

\begin{table}[h]
\centering
\small
\setlength{\tabcolsep}{10pt}
\begin{tabular*}{\columnwidth}{l@{\extracolsep{\fill}}cccc}
\toprule
 & \multicolumn{2}{c}{\textbf{Response Quality}} & & \\
\cmidrule(lr){2-3}
\textbf{Prefix} & \textbf{Good} & \textbf{Bad} & \textbf{Total} & \textbf{\% Bad} \\
\midrule
Good   & 1{,}266 & 72  & 1{,}338 & 5.4 \\
Bad    & 48      & 28  & 76      & 36.8 \\
\midrule
Total  & 1{,}314 & 100 & 1{,}414 & 7.1 \\
\bottomrule
\end{tabular*}
\caption{Contingency matrix of prefix sensibility against RelayS2S
response quality on real speech (GPT-4.1 back-end). Rows partition the
1{,}414 committed turns by whether the human annotator judged the
prefix sensible.}
\label{tab:contingency}
\end{table}

The verifier filters the incoming stream effectively: only 5.4\%
(76/1{,}414) of committed turns contain a non-sensible prefix. On the
remaining 94.6\% of turns where the prefix is sensible, RelayS2S
essentially matches the cascaded baseline (5.4\% vs.\ 4.9\% low-quality
rate). On the 76 bad-prefix turns, the slow path successfully recovers
to generate an acceptable response in 63.2\% (48/76) of cases, showing
that prefix-conditioned generation actively repairs rather than
passively propagates acoustic or semantic flaws. The remaining 1.98\%
of committed turns (28/1{,}414, where both prefix and response are bad)
represents the irreducible cost of selective handoff.

\subsection{Turn-Taking Performance}
\label{app:turntaking}

Because the fast-path S2S model is responsible for all real-time
dialogue control in RelayS2S, we separately evaluate its turn-taking
accuracy on the held-out set. For each control token we compute
precision, recall, and F1 with a tolerance window of one frame
(${\pm}160$\,ms). Table~\ref{tab:turntaking} reports the results.

\begin{table}[t]
\centering
\small
\renewcommand{\arraystretch}{1.2}
\setlength{\tabcolsep}{5pt}
\begin{tabular}{lccc}
\toprule
\textbf{Event} & \textbf{Precision} & \textbf{Recall} & \textbf{F1} \\
\midrule
Start speaking (\texttt{[BOS]}) & 78.9 & 96.1 & 86.7 \\
Stay silent (\texttt{[SIL]})    & 99.9 & 99.6 & 99.7 \\
Stop speaking (\texttt{[STP]})  & 89.2 & 97.5 & 93.2 \\
Backchannel (\texttt{[BOC]})    & 36.3 & 72.7 & 48.4 \\
\bottomrule
\end{tabular}
\caption{Turn-taking event detection performance of the fast-path S2S model,
evaluated with a ${\pm}1$ frame tolerance window.}
\label{tab:turntaking}
\end{table}

The model achieves strong performance on the three core turn-management
events: stay-silent F1 of 99.7\% indicates that the model rarely speaks
out of turn, start-speaking recall of 96.1\% shows it reliably detects
response opportunities, and stop-speaking F1 of 93.2\% confirms robust
interruption handling. Backchannel prediction is harder (F1 48.4\%),
with high recall (72.7\%) but lower precision (36.3\%), reflecting the
inherent subjectivity of when a backchannel is appropriate; we therefore treat it as auxiliary behavior rather than a primary evaluation target.

\subsection{Prefix verification Ensemble Details}
\label{app:prefix_check}
As detailed in Section \ref{sec:gating}, the prefix verification training dataset labels are constructed using a model ensemble strategy. The ensemble comprises three distinct large language models spanning varied parameter scales and architectures: Gemma-4-31B \cite{gemma4model}, Qwen3.6-27B \cite{qwen36model}, and GPT-OSS-120B \cite{gptoss120b}.

Each ensemble worker executes up to 3 retries under a strict JSON enforcement configuration using the specific rubric prompt detailed below. The final ground-truth training label is determined via a majority vote.

\begin{promptbox}[title={Verification Ensemble Prompt}]
You are evaluating whether a response PREFIX (the opening of an assistant reply) is a sensible way to begin replying to the user's latest message.

You will see:
- The conversation context.
- The PREFIX under evaluation. This may be as short as one word or as long as a couple of sentences.
- A continuation that a strong model wrote starting from that prefix, doing its best to produce a sensible reply.

The continuation is EVIDENCE about whether the prefix forced the model into a bad direction. You are NOT grading the continuation. A continuation can hallucinate, drift, or fail on its own -- that does not make the prefix bad unless the prefix itself caused or committed to the failure.

--- WHAT MAKES A PREFIX NOT SENSIBLE ---

A prefix is NOT SENSIBLE only if it CONCRETELY commits the response to a bad direction. Specifically:

1. Ungrounded stance. The prefix affirms, denies, or agrees with something the user did not actually say or establish.
   - Affirmative: "Yes" / "Yeah" / "That's right" / "Correct" / "Exactly" with no clear proposition to agree with.
   - Negative: "No" / "That's wrong" when the user did not make a claim to reject.
   - Refusal: "I can't help with that" / "I'd rather not" when the request is benign.
   - Sycophantic agreement with a user claim that is factually wrong.

2. Stated incorrect content. The prefix itself asserts a wrong fact, a fabricated source, an invented entity, or a misattribution. The wrong content must be visible in the prefix's OWN WORDS -- do not infer it from the continuation. "The Eiffel Tower, built in" has not committed to a date; "The Eiffel Tower, built in 1923," has.

3. Misread of the user. The prefix locks in a topic, intent, or framing that does not match the user's latest message. A reframe ("The underlying question here...", "Before answering...") is fine if it can plausibly land on the user's actual concern.

4. Constraint or persona violation. The user set an explicit rule (one-syllable words only, JSON only, calculator persona, specific format) and the prefix already breaks it.

5. Wrong refusal direction. The prefix refuses a benign request, OR begins to comply with a request that should be refused on safety grounds. A prefix that asks for clarification on a genuinely ambiguous or unsafe request is sensible.

--- WHAT DOES NOT MAKE A PREFIX NOT SENSIBLE ---

- The prefix is short and uncommitted ("Sure, " / "Here's " / "Let me " / "Great, " / "Okay, so "). If nothing specific has been committed to yet, default to SENSIBLE.
- The continuation hallucinates a fact, cites a fake source, or wanders off topic, but nothing in the prefix forced that. Attribute the failure to the continuation, not the prefix.
- The opener is stylistically bland, hedgy, or could be better written. Sensibility is about not foreclosing a good reply, not about being the best possible opener.
- The prefix is grammatically incomplete. A good continuation can finish the sentence.

--- HOW TO DECIDE ---

Ask: could a careful, well-informed continuation have followed this prefix and produced a correct, on-topic, appropriately-toned reply?
- If yes -> SENSIBLE, even if the continuation you actually see failed.
- If no, because the prefix itself states something wrong, agrees with nothing, refuses in the wrong direction, or violates a constraint -> NOT SENSIBLE.

When uncertain, default to SENSIBLE. Reserve NOT SENSIBLE for cases where you can point to specific text in the prefix that does the committing.

-----
Conversation:
{{ context }}
-----
PREFIX (under evaluation): "{{ prefix }}"
-----
Continuation that followed: {{ continuation }}
-----

Reply with valid JSON only, in this exact shape:
{"sensible": true|false}
\end{promptbox}

\subsection{Slow-Path Prompt Templates}
\label{app:response_continuation}
\subsubsection{Prefix-Conditioned Generation and Repair}
When the prefix verifier commits a fast-path draft, the downstream slow-path LLM is invoked to complete the response. The generation is forced to decode directly from the committed token history and is explicitly instructed to self-correct any flaws that leaked through the fast path.

\begin{promptbox}[title={Prefix-Conditioned Generation Prompt}]
You are continuing a conversation in a real-time dialogue system.

### CONVERSATION HISTORY:
{{ history_str }}

### TASK:
The Assistant has already started speaking their next response. The initial part of the response (the prefix) has already been played aloud to the user and cannot be taken back. Your task is to continue the response starting from the exact point where the prefix ends.

### CORRECTION AND CONTINUATION GUIDELINES:
1. Assume the prefix has already been heard by the user exactly as it is written.
2. If the prefix is contextually appropriate and correct, continue it smoothly to complete the response naturally.
3. If the prefix contains a factual error, hallucination, repetition, or is contextually inappropriate, you must perform a graceful self-correction or conversational repair within the continuation (e.g., by saying "... wait, I mean...", "... actually, let me correct that...", or pivoting to the right answer naturally) without breaking the flow.

### OUTPUT FORMAT:
Output ONLY the continuation of the response. Do not repeat the prefix. Do not include labels, markdown conversational fillers, or any meta-text.

### PREFIX TO CONTINUE:
{{ forced_prefix }}
\end{promptbox}

\subsubsection{Standard Fallback Generation}
\label{app:response_continuation_standard}
On verifier fallback, the system relies on unconstrained response generation from the back-end cascade pipeline utilizing the following prompt template:

\begin{promptbox}[title={Standard Fallback Generation Prompt}]
You are generating a response in a real-time dialogue system.

### CONVERSATION HISTORY:
{{ history_str }}

### TASK:
Generate the Assistant's next response based on the conversation history above. The response will be played aloud to the user in real time, so it should sound natural when spoken.

### RESPONSE GUIDELINES:
1. Respond directly and naturally to the user's most recent message.
2. Ensure the response is factually accurate, contextually appropriate, and consistent with the tone and style established in the conversation history.
3. Avoid hallucinations, unnecessary repetition, or filler content.
4. Keep the response conversational and suitable for being spoken aloud.

### OUTPUT FORMAT:
Output ONLY the Assistant's response. Do not include labels, markdown, conversational fillers, or any meta-text.
\end{promptbox}

\subsection{Real Voice Filtering Criteria}
\label{app:real_voice}
To filter uncooperative, incomplete, or purely narrative text from raw ASR corpora, we employ Gemini-3-Flash. The configuration leverages explicit criteria to retain only self-contained prompts suitable for an AI assistant:

\begin{promptbox}[title={Real Voice Filtering Prompt}]
Keep ASR utterances that are BOTH (1) fully context-independent and
(2) plausibly something a user would say to an AI assistant.

CORE TEST - reject if any is no:
(a) Are all WHO/WHAT/WHEN/WHICH clear from the sentence alone?
(b) Does it read like a real utterance (question, request, opinion,
    standalone fact) - not a mid-article fragment?
(c) If addressed to "you," could it sensibly target an AI (which has
    no location, family, age, job, hometown, hobbies, feelings, or
    personal history)?

REJECT - personal probes at the listener
Questions/remarks about the listener's life, identity, relationships,
or experiences. Nonsensical when aimed at an AI.
  - Origin/status: "Where are you from?", "Are you married?"
  - Greetings-as-questions: "How are you doing?"
  - Personal plans/reactions: "Where are you going for spring break?",
    "Drums are awesome."
  - Tone probes: "Are you being sarcastic?"
  - Opinion solicitations as turn-taking: "...do you?"
(Generic AI-answerable questions like "What is the capital of France?"
are fine.)

REJECT -- context-dependence
  - Unresolved pronouns (he/she/it/they/his/her) unless antecedent is
    a globally famous entity named in-sentence.
    Reject: "Chris Hattingh filled his seat."
  - Bare names (first or last alone, or full names of obscure people)
    as subjects/possessives. Single-name refs are NEVER globally
    identifiable.
    Reject: "Navarro is fluent in Spanish, French and English."
    Accept: "Einstein's theory of general relativity..."
  - Unresolved demonstratives (this/that/these/those/there).
    Reject: "In this sense, Jesus becomes a God-Man."
  - Discourse connectives implying prior context: then, also, instead,
    too, however, moreover, another, finally, for the first time.
    Reject: "PowerShell uses backtick instead."
  - Definite "the X" where X isn't specified in-sentence. Allow if
    fully specified ("the capital of France") or unambiguous generic
    ("the immune system").
    Reject: "The park is featured in an episode of K-On!."
  - Fragments, headlines, titles, isolated quotes, maxims, captions.
    Reject: "Delegata potestas non potest delegari."

REJECT -- encyclopedia narration
Even if referents resolve, reject Wikipedia-style prose:
  - Dry biographical reporting:
    "John Smith was born in 1842 in a small Yorkshire village."
  - Infobox stat-dumps, transit/schedule/operator facts:
    "The 1998 Honda Civic has a 1.6-liter engine and 106 horsepower."
  - Plot/episode summaries, credits-style sentences.
  - "X is a [type] in/at [location]" landmark/institution openers:
    "Breed's Hill is a glacial drumlin in Charlestown, Boston."

ACCEPT -- self-contained AND prompt-like
  - AI-answerable questions: "What is the boiling point of mercury?"
  - Requests: "Write a haiku about autumn."
  - First-person opinions: "I think Mondays are the worst day."
  - Generic claims: "Photosynthesis converts light into chemical energy."
  - Facts about globally identifiable entities:
    "The Eiffel Tower was completed in 1889."

When in doubt, REJECT.

Output one line of JSON only:
{"is_standalone_prompt": true|false}

Transcription:
"""
{{ transcription }}
"""
\end{promptbox}

\subsection{Evaluation Rubric for Textual Response Quality}
\label{app:textual_quality}
\begin{promptbox}[title={Evaluation Rubric Prompt}]
You are an expert AI quality auditor. Score the assistant's LAST RESPONSE based on the provided CONTEXT.

SCOPE:
- Grade ONLY the assistant's last response. Earlier turns are context, not subjects.
- If the user's request is ambiguous or unanswerable, a graceful clarification or
  honest refusal can score up to 5, provided it is accurate and well-phrased.

DIMENSIONS (consider all four; do not output them separately):
1. Relevance -- directly addresses the request without drifting off-topic.
2. Helpfulness -- provides useful, complete, actionable value for the request.
3. Accuracy -- free of hallucinations, contradictions, invented details, overstatements.
4. Clarity -- well-organized and easy to follow in the stated modality.

HARMFUL CONTENT OVERRIDE:
If the response contains content that is harmful, dangerous, or unsafe (and the
request did not justify it), score = 1 regardless of other dimensions.

ACCURACY CAP RULES (apply after the harm check):
- CLEAN: no factual mistakes anywhere. No cap; can reach 5.
- SELF-CORRECTED: a mistake appears earlier in the response but is explicitly
  repaired later, leaving the listener with the right answer. Cap = 4.
  Earlier, smoother repairs sit near 4; late or clumsy repairs sit lower.
- UNCORRECTED: a factual error, hallucination, or misleading claim remains
  standing or is glossed over without acknowledgement. Cap = 3.

AGGREGATION INTO A SINGLE SCORE (1 to 5 integer):
Start from the accuracy cap, then adjust DOWN for weaknesses in relevance,
helpfulness, or clarity. Do not adjust UP past the cap.

5 = Excellent. Fully addresses the request; accurate; clear; no mistakes
    that required repair. Relevance, helpfulness, and clarity all strong.
4 = Good. Mostly complete, relevant, and accurate, with small gaps or minor
    clarity issues. OR: the ceiling for a smooth, early self-correction.
3 = Fair. Surface-level answer with minor uncorrected issues, OR a clumsy
    self-correction that left the listener confused even though the final
    answer was right, OR notable gaps in helpfulness/relevance.
2 = Poor. Partially relevant but missing key content, unclear, or contains
    notable uncorrected errors that affect the takeaway.
1 = Very poor. Fails the request, materially inaccurate, confusing, or harmful.

OUTPUT FORMAT:
Return ONLY this JSON, with no surrounding prose, no code fences, no commentary:
{"score": <integer 1-5>}

--------------------
CONTEXT:
{{ context }}
--------------------
RESPONSE (to be graded):
Assistant: {{ response }}
\end{promptbox}

\end{document}